\documentclass[conference]{IEEEtran}
\usepackage{times}

\usepackage[numbers]{natbib}
\usepackage{multicol}

\usepackage{graphicx} % for pdf, bitmapped graphics files
\usepackage{mathptmx} % assumes new font selection scheme installed
\usepackage{amsmath} % assumes amsmath package installed
\usepackage{amssymb}  % assumes amsmath package installed
\usepackage{tikz}
\usepackage{refstyle}
\usepackage{caption}
\usepackage{subcaption}
\usepackage{amssymb}
\usepackage{array}
\usepackage{color,soul}
\usepackage{breqn}
\usepackage{multirow}
\usepackage[mathscr]{euscript}
\usepackage[colorlinks=true,linkcolor=black,anchorcolor=black,citecolor=black,filecolor=black,menucolor=black,runcolor=black,urlcolor=black]{hyperref}

\begin{document}

% paper title
\title{Towards Monocular Vision based Obstacle Avoidance through Deep Reinforcement Learning}

\author{\authorblockN{Linhai Xie, Sen Wang, Andrew Markham and Niki Trigoni}
\authorblockA{Department of Computer Science, University of Oxford,
Oxford OX1 3QD, United Kingdom\\
\tt\small \{firstname.lastname\} @cs.ox.ac.uk}}

\maketitle

\begin{abstract}
Obstacle avoidance is a fundamental requirement for autonomous robots which operate in, and interact with, the real world. When perception is limited to monocular vision avoiding collision becomes significantly more challenging due to the lack of 3D information. Conventional path planners for obstacle avoidance require tuning a number of parameters and do not have the ability to directly benefit from large datasets and continuous use. In this paper, a dueling architecture based deep double-Q network (D3QN) is proposed for obstacle avoidance, using only monocular RGB vision. Based on the dueling and double-Q mechanisms, D3QN can efficiently learn how to avoid obstacles in a simulator even with very noisy depth information predicted from RGB image. Extensive experiments show that D3QN enables twofold acceleration on learning compared with a normal deep Q network and the models trained solely in virtual environments can be directly transferred to real robots, generalizing well to various new environments with previously unseen dynamic objects. 
\end{abstract}

\IEEEpeerreviewmaketitle

\section{Introduction}

% RGB image + Deep Reinforcement Learning + fast learning + directly from simulator to real world

When mobile robots operate in the real world, subject to ever varying conditions, one of the fundamental capabilities they need is to be able to avoid obstacles. A long established problem in robotics, obstacle avoidance is typically tackled by approaches based on ranging sensors \cite{everett1989survey}, e.g. laser scanner and sonar. However, ranging sensors only capture limited information and some of them are expensive or are too heavy/power consuming for a particular platform e.g. a UAV. Monocular cameras on the other hand, provide rich information about the robot's operating environments, are low-cost, light-weight and applicable for a wide range of platforms. However, when perception of range is obtained by monocular vision, i.e., RGB imagery, the obstacle avoidance problem becomes surprisingly difficult. This is because the 3-D world is flattened into a 2-D image plane, eliminating direct correspondence between pixels and distances.

% Non-Learning based Approaches
A standard framework to solve this problem consists of two steps, the first of which utilizes visual information to infer traversable spaces and obstacles, and then secondly applying conventional path planning strategies. Recovering visual geometry is a common approach to detecting obstacles, e.g. through optical flow \cite{souhila2007optical,mccarthy2004performance}, detection of vanishing points \cite{bills2011autonomous} and even visual SLAM \cite{mur2015orb}. Segmenting traversable areas, such as floors, based on visual appearance \cite{ulrich2000appearance} is also a popular method. Once the surroundings are understood, various conventional path planners can then be employed to drive robots along traversable routes \cite{fox1997dynamic}. Although the described methods are able to decouple planning from visual information and hence benefit from conventional path planners, they usually require a large number of parameters which need to be manually tuned to plan feasible paths. It is also challenging for them to automatically adapt to the new operating areas.

% Supervised learning
Deep learning nowadays has shown its great performance in robotics and computer vision\cite{wang2017deepvo,clark_cvpr16,Clark_AAAI17}. And supervised deep learning based path planning which learns how to avoid collision is becoming increasingly popular. In particular, with the recent advances of deep learning, several end-to-end supervised learning approaches are proposed to directly predict control policies from raw images \cite{kim2015deep,giusti2016machine,tai2016deep} without following the previous two-step framework. Therefore, they can avoid complex modeling and parameter tuning of conventional path planners. Convolutional Neural Networks (CNNs), for example, are trained to enable flying robots to navigate in complex forest environments in \cite{giusti2016machine}. However, due to their supervised nature, these approaches need manual labeling which is time-consuming and labor-intensive to obtain.

% Self-Supervised Learning
Self-supervised learning can be used to automatically generate labels for path planners with the aid of additional feedback. For instance, in \cite{yang2017obstacle} an algorithm is designed to label trajectory classes for a CNN based model by using 3D cloud points. \citet{gandhi2017learning} proposes to train a drone controller by predicting when it crashes. Although self-supervised learning is a feasible approach to benefiting from large dataset without human supervision, the learnt policy is essentially bounded by the label generating strategy.  

\begin{figure*}
	\centering
	\includegraphics[width=0.85\textwidth]{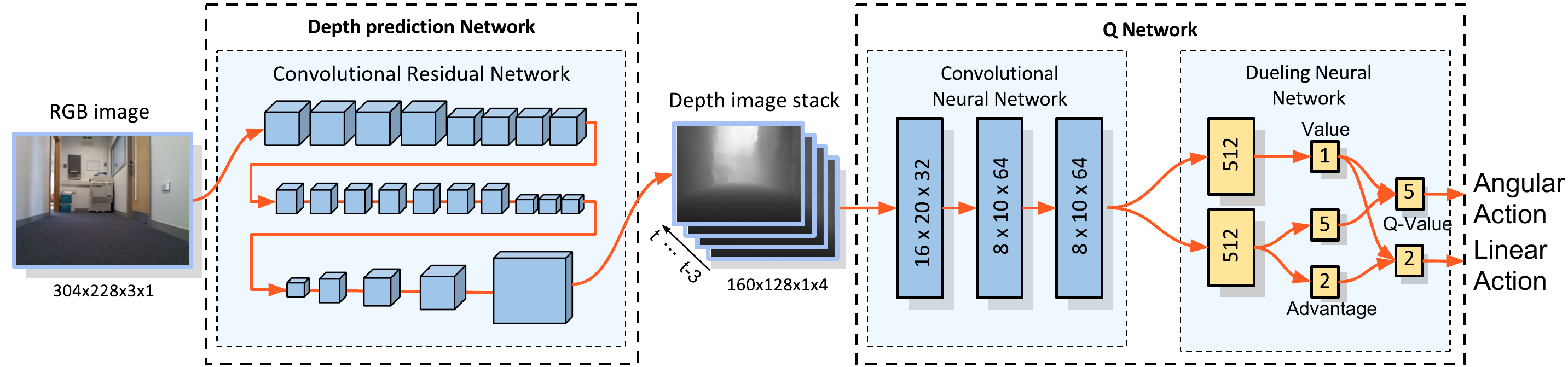}
	\caption{Network architecture of monocular image based obstacle avoidance through deep reinforcement learning. A fully convolutional neural network is firstly constructed to predict depth from a raw RGB image. It is then followed by a deep Q network which consists of a convolutional network and a dueling network to predict the Q-value of angular actions and linear actions in parallel.}
	\label{fig:model}
\end{figure*}

% Reinforcement Learning based Approaches
% Zhu et al. \cite{zhu2017icra} proposed a target driven visual navigation method based on an actor-critic network. By leveraging high quality simulations, the learned policy can be adapted to real world. 
Reinforcement learning explores policies though trials, and has been applied to vision based obstacle avoidance in \cite{michels2005high}. However, the raw image is encoded as several levels of depth to predict a suitable control strategy. Deep reinforcement learning (DRL) has recently been shown to achieve superhuman performance on games by fully exploring raw images \cite{mnih2015human}. Since DLR usually utilizes a much weaker learning signal compared with supervised learning, it requires a much larger training dataset. This makes it difficult to directly use DLR for robotic applications in reality. Therefore, simulations which have a failure-and-recovery mechanism are usually used for training rather than real world exploration \cite{Mirowski2017a}. The trained networks can then be transferred to real robots. Although this has been successful by using laser scanner \cite{tai2017virtual} and depth images \cite{tai2016towards}, it is significantly more difficult for vision based models \cite{gandhi2017learning}. Recently \citet{sadeghi2016rl} propose to train a network as a collision predictor entirely in a 3D CAD model simulator and highly randomize the rendering settings, approximately regarding the real world as a subset of training data. Although their model can be extended into real world, it requires substantial computational resources to generate the huge dataset and train it.

In this paper, we focus on the problem of obstacle avoidance with monocular camera. More specifically, our contributions are:
\begin{itemize}
  \item A two-phase deep neural network is proposed to achieve monocular vision based obstacle avoidance.
  \item Dueling architecture based deep double Q network (D3QN) is applied to obtain a high speed for end-to-end learning with limited computational resources for obstacle avoidance task.
  \item The knowledge learnt from simulation can be seamlessly transfered to new scenarios in the real world.
  \item Extensive real-world experiments are conducted to show the high performance of our network.
\end{itemize}

The rest of this paper is organized as follows. The proposed D3QN model is described in Section II. Experimental results are given in Section III, followed by conclusions drawn in Section IV.

\section{Deep Q Network for Monocular Vision based Obstacle Avoidance}

Since deep Q network (DQN) has been shown to be trainable directly benefit from raw images \cite{mnih2015human}, most DQN models used for obstacle avoidance are based on this version \cite{tai2016towards,zhang2016vision}. Although this architecture can eventually achieve reasonable results, it tends to overestimate Q values and takes a long time to train as discussed in \cite{van2016deep,DBLP:journals/corr/WangFL15}. This leads to intensive computational resources for training in simulators. In this section, an advanced architecture, D3QN, is introduced to boost both performance and training efficiency for monocular vision based obstacle avoidance. 

\subsection{Problem Definition}

The monocular vision based obstacle avoidance problem can be considered as a decision making process where the robot is interacting with environments with a monocular camera. The robot chooses an action $a_t \in \mathscr{A}$ according to the camera image $x_t$ at time $t \in [0,T]$, observes a reward signal $r_t$ produced by an assessor (reward function) and then transits to the next observation $x_{t+1}$. The aim of the algorithm is to maximize the accumulative future reward $R_t = \sum_{\tau=t}^{T}\gamma^{\tau-t}r_\tau$, where $\gamma$ is the discount factor. 

Given the policy $a_t = \pi(x_t)$, the action-value (Q-value) of a state-action pair $(x_t, a_t)$ can be defined as follows

\begin{equation}
  \begin{aligned}
    Q^\pi(x_t,a_t) &= \mathbb{E}[R_t|x_t,a_t,\pi],\\
  \end{aligned}
\end{equation}
The Q-value function can be computed using the Bellman equation

\begin{equation*}
	Q^\pi(x_t,a_t) = \mathbb{E}[r_t+\gamma\mathbb{E}[Q^\pi(x_{t+1},a_{t+1})|x_t,a_t,\pi].
\end{equation*}
By choosing the optimal action each time where $Q^*(x_t,a_t)=max_\pi\mathbb{E}[R_t|x_t,a_t,\pi]$, we can have the optimal Q-value function

\begin{equation}
Q^*(x_t,a_t)=\mathbb{E}_{x_{t+1}}[r+\gamma \max\limits_{a_{t+1}}Q^*(x_{t+1},a_{t+1})|x_t,a_t],
\end{equation}
which indicates that the optimal Q-value we can obtain at time $t$ is the current reward $r_t$ plus the discounted optimal Q-value available at time $t+1$. Rather than computing the 
Q-value function directly over a large state space, the problem can be solved by approximating this optimal Q-value function with deep neural networks, which is the main principle behind DQN.

\subsection{Dueling Network and Double Q-Network}\label{sec:dueling and double network}

\begin{figure}
  \centering
  \includegraphics[width=0.8\linewidth]{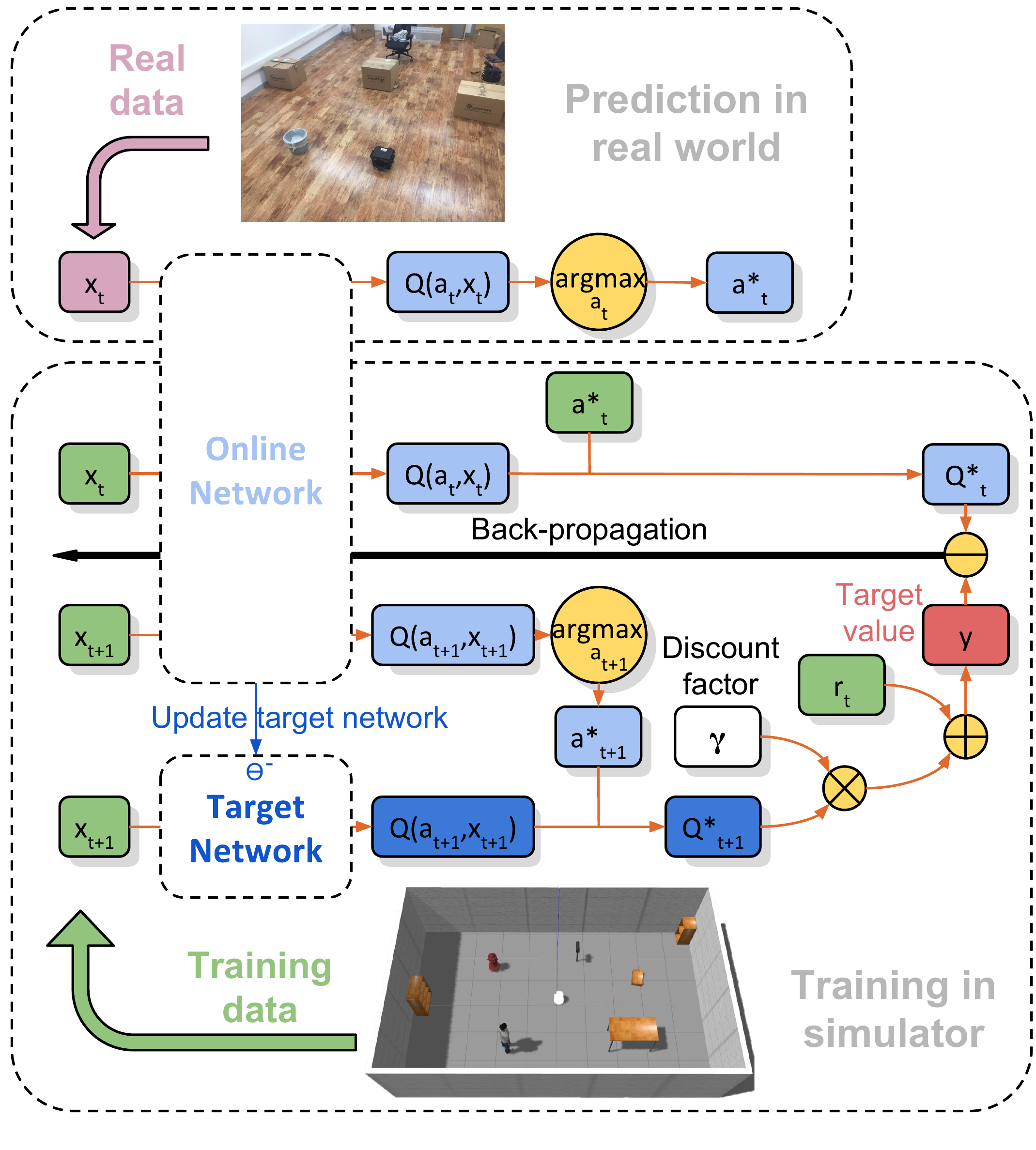}
  \caption{When given a batch of training data, including current state $x_t$, action $a$, reward $r$, and resulting state $x_{t+1}$, the training procedure of D3QN is shown in the figure. $\oplus$, $\ominus$ and $\otimes$ are element-wise operation for addition, subtraction and multiplication.}
  \label{fig:training}
\end{figure}

With the intuition that it is unnecessary for all actions to be estimated at each state $s$, \citet{DBLP:journals/corr/WangFL15} propose the dueling network architecture. In traditional DQN only one stream of fully connected layers is constructed after the convolution layers to estimate the Q-value of each action-state pair, given the current state. However, in the dueling network, two streams of fully connected layers are built to compute the value and advantage functions separately, which are finally combined together for computing Q-values. This two-stream dueling network structure is shown in the last section of Fig.\ref{fig:model}. It has demonstrated a large improvement either on performance or training speed in a number of ATARI games (but not all). Thus, it is exploited in our model to facilitate the learning of obstacle avoidance.

The prototype DQN in \cite{mnih2015human} utilizes a target network alongside an online network to stabilize the overall network performance. The target network is a duplicate of the online one. However, unlike the online network which updates weights by back-propagation at every training step, the weights of the target network are fixed over a short period and then copied from online network. Based on this two-network framework, \citet{van2016deep} claim that the online network should be used to select actions while the target network should be used solving the problem of overoptimistic value estimation \cite{hasselt2010double}. This procedure is shown in Fig.\ref{fig:training}. More specifically, the resulting state $x_{t+1}$ is employed by both the online and target network to compute the optimal value ${Q'}^*$ for time $t+1$. Then, with the discount factor $\gamma$ and current reward $r_t$, the target value $y$ at $t$ is obtained. Finally, the error is calculated by subtracting the target value with the optimal value $Q^*$ predicted by the online network, given current state $x$, and is then back-propagated to update the weights.

With these two techniques, the proposed D3QN is expected to be more data efficient to speed up learning. We will discuss the details of the model architecture in Section \ref{sec: model and training settings}.

\subsection{From Appearance to Geometry}

Since DRL needs huge amounts of data and time to train, its performance is usually demonstrated in simulated environments. In order to apply it in practice for robotic applications, a feasible solution is to train the models in simulator and then transfer them to real robots. However, this is highly challenging for vision based techniques due to the significant differences between virtual and real environments due to appearance, illumination, etc. To solve this problem, we propose to derive a geometric representation from the RGB imagery. 

\begin{figure}
	\centering
	\begin{subfigure}{0.31\linewidth}
		\centering
		\includegraphics[width=\linewidth]{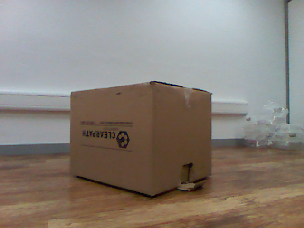}
		\caption{RGB}
	\end{subfigure}
	\begin{subfigure}{0.29\linewidth}
		\centering
		\includegraphics[width=\linewidth]{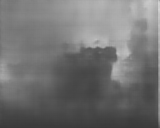}
		\caption{Predicted Depth}
	\end{subfigure}
	\begin{subfigure}{0.29\linewidth}
		\centering
		\includegraphics[width=\linewidth]{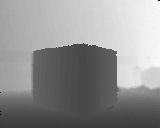}
		\caption{Kinect's Depth}
	\end{subfigure}
	\caption{Images of RGB, predicted depth and Kinect's depth. Note the noisy depth predicted from the network.}
	\label{fig:DepthComp}
\end{figure}

As shown in Fig.\ref{fig:model}, the first part of the D3QN model is inspired by a fully convolutional residual network (FCRN) in \cite{laina2016deeper}, predicting depth information from a single RGB image. However, as the depth image used is estimated by a deep neural network rather than obtained from a 3D sensor, e.g., Kinect, they are very inaccurate in practice, see \figurename{\ref{fig:DepthComp}}. This makes it impossible to directly use traditional ranging sensor based methods for obstacle avoidance.

In order to tackle this serious challenge, the depth images used for training in the simulator are corrupted with random noise and image blur. We found this is critical to ensure the trained models are transferable from simulation to reality, and generalize well in real world. 

\subsection{Model and Training Settings}\label{sec: model and training settings}

The D3QN model is built based on the dueling and double techniques. Its architecture is shown in Fig.\ref{fig:model} and corresponding parameters are given in Table \ref{tab: Q network structure}. Specifically, it has three convolutional layers, specified with filter size (height, width, channel), and three fully connected layers for two streams of dueling architecture discussed in \ref{sec:dueling and double network}.

\begin{table}
\centering
\caption{Parameters of D3QN Model for Obstacle Avoidance}
\label{tab: Q network structure}
\begin{tabular}{c|cc}
Name of layer      & \begin{tabular}[c]{@{}c@{}}Size of filters or \\ number of neurons\end{tabular} & Stride \\ \hline
Conv 1             & (10, 14, 32)                                                                    & 8      \\
Conv 2             & (4, 4, 64)                                                                      & 2      \\
Conv 3             & (3, 3, 64)                                                                      & 1      \\
FC 1 for advantage & 512                                                                             & -      \\
FC 1 for value     & 512                                                                             & -      \\
FC 2 for advantage of angular actions & 5                                                            & -      \\
FC 2 for advantage of linear actions & 2                                                             & -      \\
FC 2 for value     & 1                                                                               & -     
\end{tabular}
\end{table}

To train the network to produce feasible control policy, robot actions need to be properly defined. Instead of the simple commands e.g. ``go ahead'', ``turn left or right'', the actions in our network are defined to control the linear and angular velocities separately in a discretised format. 

The instantaneous reward function is defined as $r = v*cos(\omega)*\delta t$ where $v$ and $\omega$ are local linear and angular velocity respectively and $\delta t$ is the time for each training loop which is set to $0.2$ second. The reward function is designed to let the robot run as fast as possible and be penalized by simply rotating on the spot. The total episode reward is the accumulation of instantaneous rewards of all steps within an episode. If a collision is detected, the episode terminates immediately with an additional punishment of $-10$. Otherwise, the episode lasts until it reaches the maximum number of steps (500 steps in our experiments) and terminates with no punishment.

\begin{figure}
  \centering
  \begin{subfigure}{.35\linewidth}
      \centering
      \includegraphics[width=\linewidth]{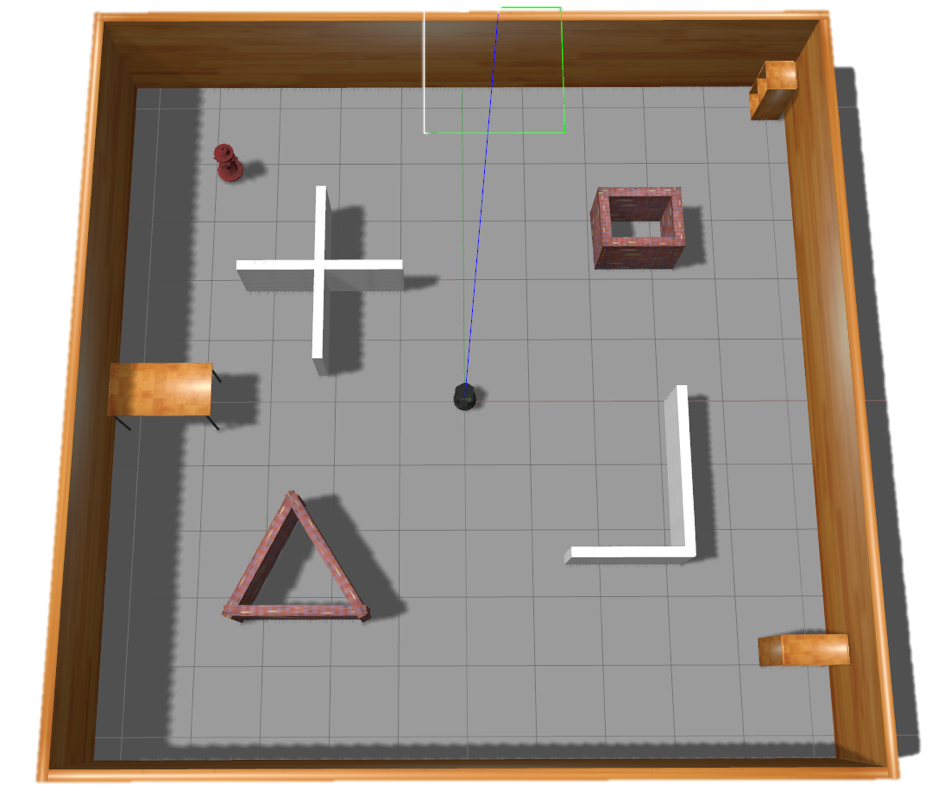}
      \caption{Simple}
  \end{subfigure}
  \quad
  \begin{subfigure}{.4\linewidth}
      \centering
      \includegraphics[width=\linewidth]{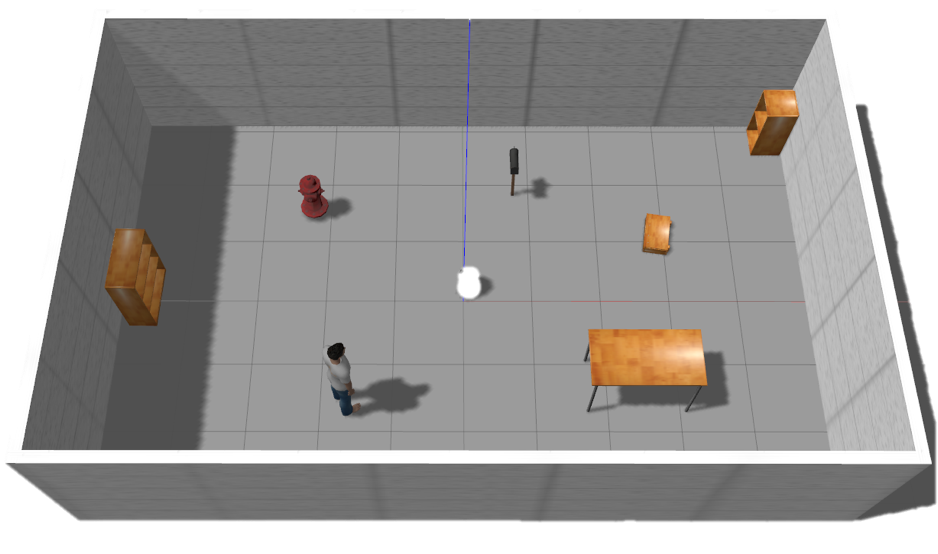}
      \caption{Complex}
  \end{subfigure}
  \caption{Two simulation worlds in Gazebo used for training.}
  \label{Fig:simulator}
\end{figure}

\section{Experimental Results}

In this section, the proposed D3QN model is evaluated in different environments. Two simulation environments, simple and complex ones, are built in Gazebo simulator for training, see Fig.\ref{Fig:simulator}. The D3QN model is firstly trained in the simple environment before being further trained in the complex scenario. The trained model in the simulator is directly tested in several different real world scenarios. The linear velocity is set to be $0.4$ or $0.2$ m/s, while the angular velocity is $\frac{\pi}{6}$, $\frac{\pi}{12}$, $0$, $-\frac{\pi}{12}$ or $-\frac{\pi}{6}$ rad/s, producing ten different behaviors. Throughout our experiments, a NVIDIA TitanX GPU is used for training while a laptop equipped with a NVIDIA GTX 860 GPU is used for real-time inference and testing in reality. The learning rate is set to $10^{-5}$ in an Adam optimizer \cite{kingma2014adam}. A Turtlebot robot is used to test the control strategy in real-time. 

\subsection{Training Efficiency with Different Models}

\begin{figure}
  \centering
  \includegraphics[width=0.7\linewidth]{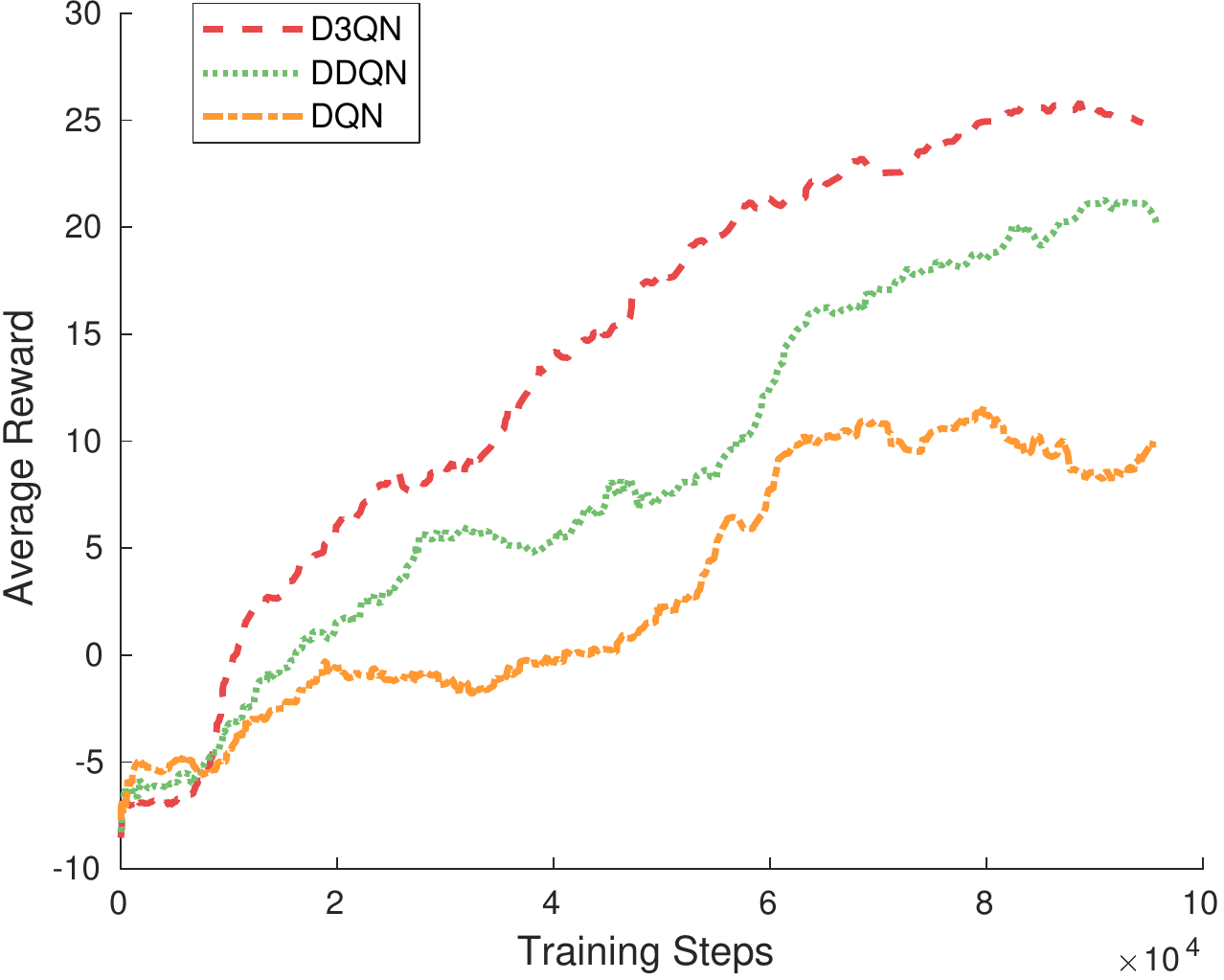}
  \caption{Smoothed learning curves of the three models with average rewards acquired by robot.}
  \label{fig:training_curves}
\end{figure}

To analyse the training efficiency and performance of the D3QN model and the advantage of introducing dueling and double techniques for obstacle avoidance, deep double Q network (DDQN) and DQN are compared. As shown in Fig.\ref{fig:training_curves}, D3QN model outperforms the others two both on the training speed and performance. Unlike DQN whose average reward only reaches 10, networks with a double network structure learn policies with higher rewards. This may be because, for the obstacle avoidance problem, the overestimation of Q value is not a problem that can be alleviated by getting more exploration. Conversely, with a longer training period, it might be more severe, preventing DQN from obtaining high performance. Therefore, the D3QN architecture is about two times faster on training than the widely used normal DQN, which not only demonstrates its appealing performance on obstacle avoidance but also suggests an interesting direction of applying it on other robotic applications. 
  
\subsection{Real World Tests}

Several experiments are conducted to directly test the trained models in real world. 

\subsubsection{Action Prediction from Static Images}

\begin{figure}
\centering
  \begin{subfigure}{0.8\linewidth}
  \centering
  \includegraphics[width=\linewidth]{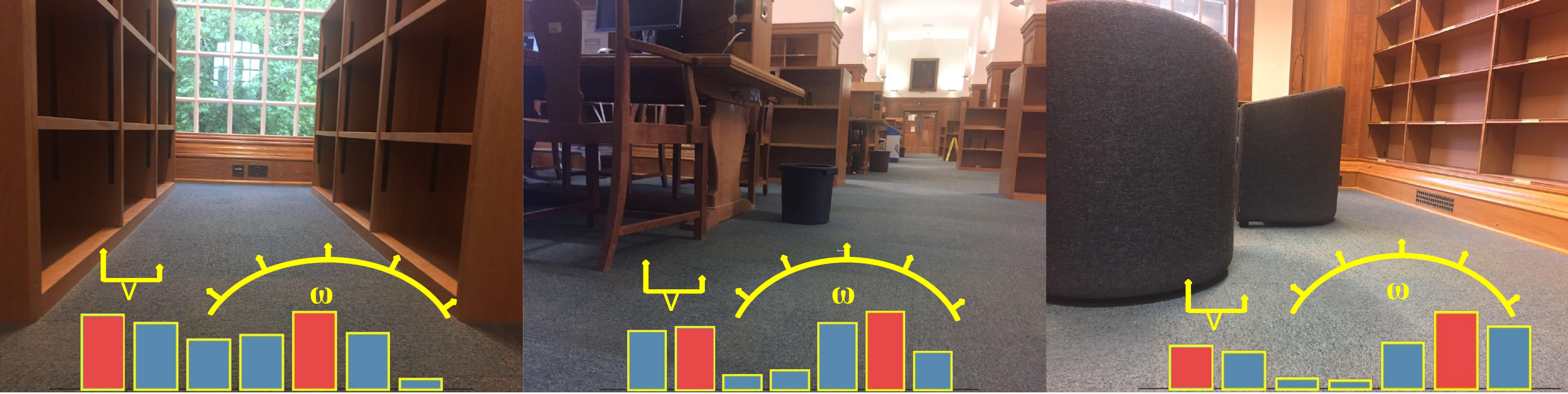}
  \end{subfigure}
  \begin{subfigure}{0.8\linewidth}
  \centering
  \includegraphics[width=\linewidth]{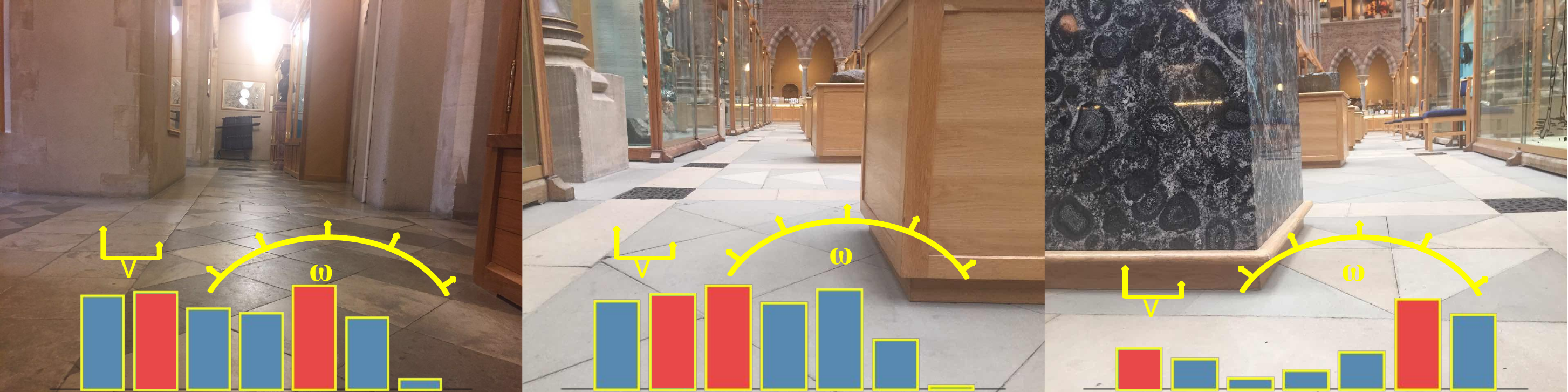}
  \end{subfigure}
  \begin{subfigure}{0.8\linewidth}
  \centering
  \includegraphics[width=\linewidth]{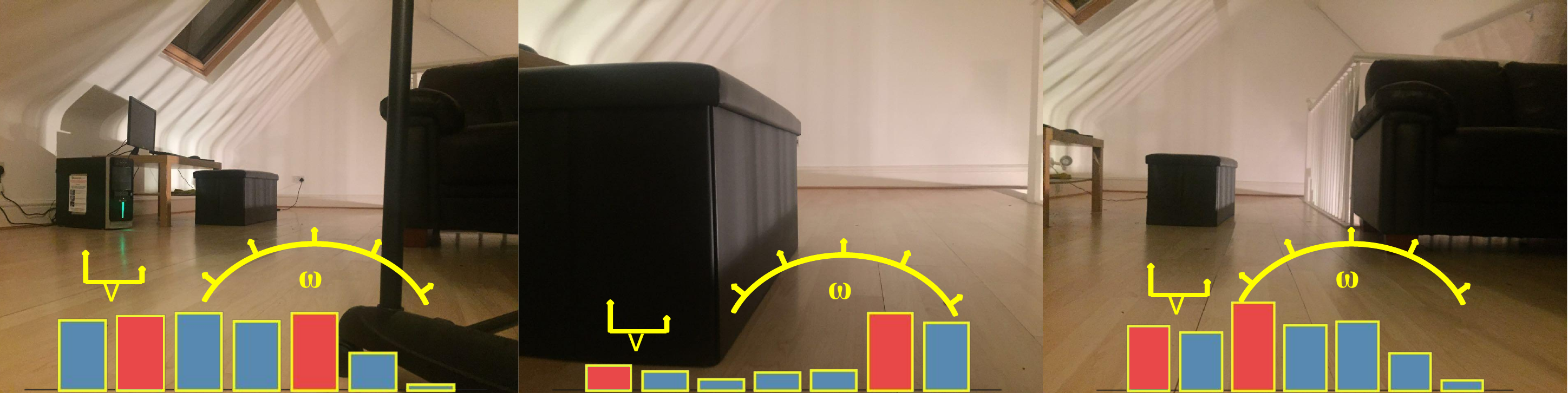}
  \end{subfigure}
  \begin{subfigure}{0.8\linewidth}
  \centering
  \includegraphics[width=\linewidth]{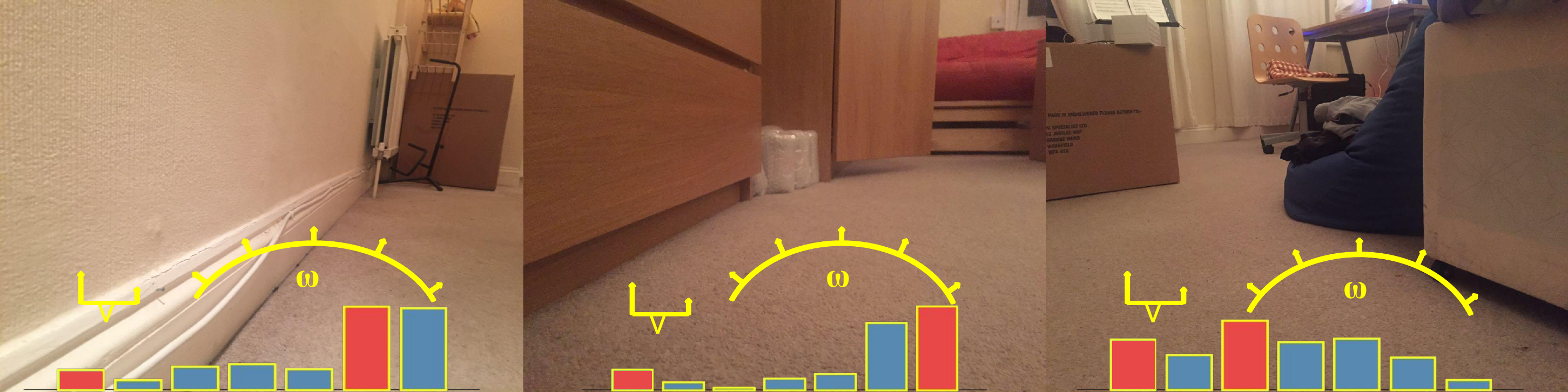}
  \end{subfigure}
\caption{Experiments in different indoor environments, e.g. library, museum, attic and bedroom (from top to bottom). The underlying bars demonstrates the Q value for each linear and angular action predicted by network, where the red ones indicate the actions greedily selected by network. Notice that the first two are for linear speed actions while the rest are for steering actions.}
\label{fig:single image test}
\end{figure}

Firstly, we examine whether for arbitrary, complex scenarios, the network is able to predict a reasonable action that will avoid obstacles. As shown in Fig.\ref{fig:single image test}, a number of RGB images taken by a hand-held camera in a variety of environments including library, museum, attic and bedroom are used to predict actions. The bars in the figure indicate the Q value of each action: the first two bars are for linear velocity $0.2$m/s and $0.4$m/s, while the rest are for the five angular velocity $\frac{\pi}{6}$rad/s, $\frac{\pi}{12}$rad/s, $0$rad/s, $-\frac{\pi}{12}$rad/s and  $-\frac{\pi}{6}$rad/s. Note that these scenarios are more complicated than the simulation ones used for training and none of them has been ``seen'' by the model before. It can be seen that the trained D3QN model is capable of producing reasonable actions to drive the robot according to the estimated Q values.

\begin{figure}
\centering
  \begin{subfigure}{\linewidth}
  \centering
  \includegraphics[width=0.9\linewidth]{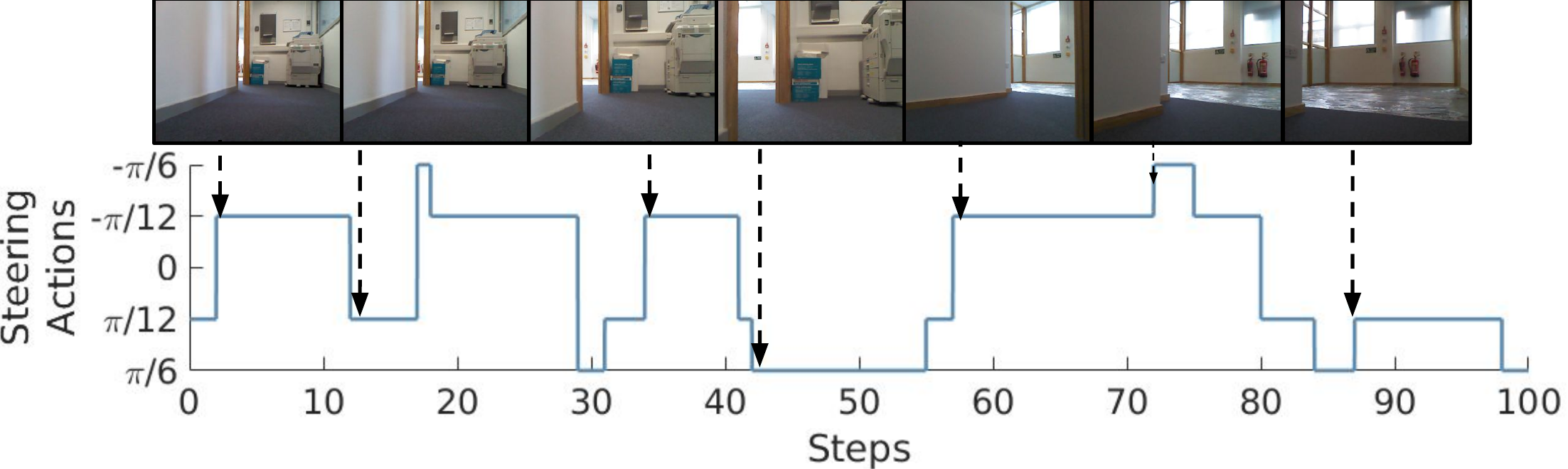}
  \caption{Passing a doorway.}
  \label{fig:passing a gate}
  \end{subfigure}
  \begin{subfigure}{\linewidth}
  \centering
  \includegraphics[width=0.9\linewidth]{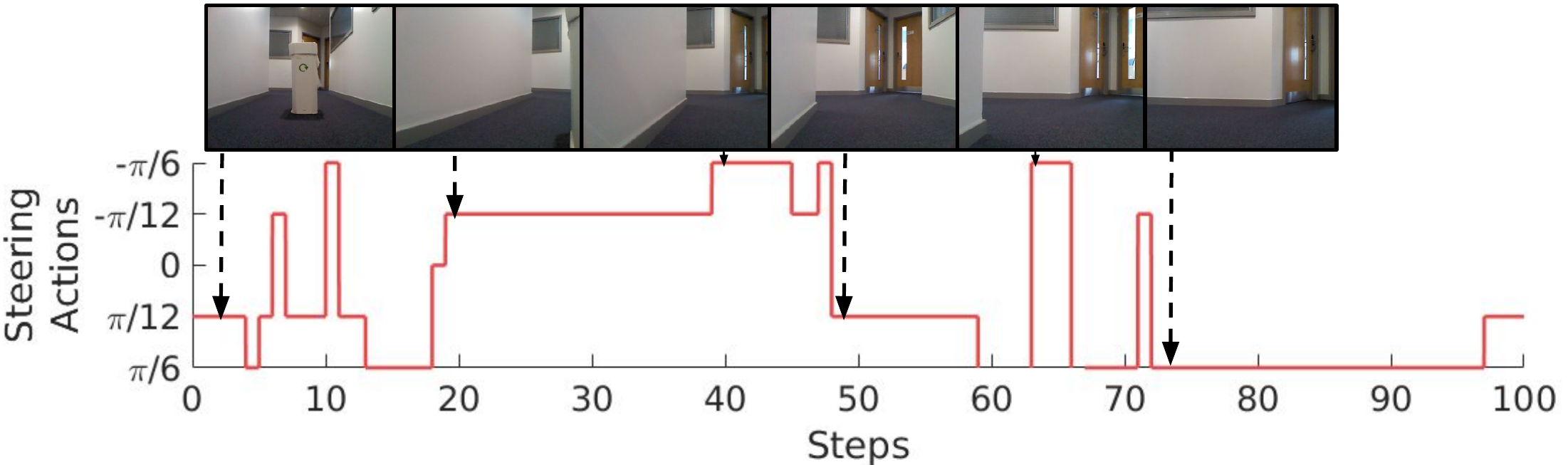}
  \caption{Driving through a corridor.}
  \label{fig:driving through a corridor}
  \end{subfigure}
  \begin{subfigure}{\linewidth}
  \centering
  \includegraphics[width=0.9\linewidth]{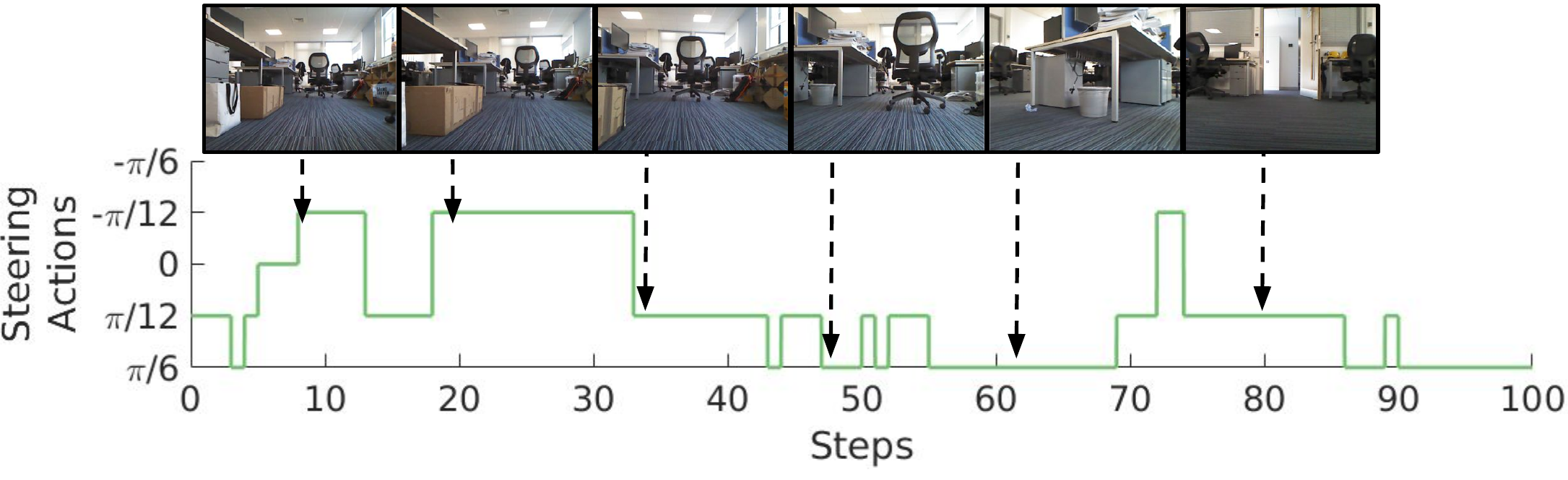}
  \caption{In an office room.}
  \label{In an office room}
  \end{subfigure}
  \caption{Real world tests in three different scenarios. The curve below the image streams shows the steering actions selected by robots at each step.}
  \label{Fig:real_test}
\end{figure}

\subsubsection{Tests in Three Different Scenarios}

The trained D3QN model is tested for short-term (20s) navigation in three different scenarios including a doorway, a corridor and an office. The steering actions and some sample images of the three tests are given in Fig.\ref{Fig:real_test}. Specifically, \figurename{\ref{fig:passing a gate}} shows the procedure of the robot passing the doorway. Although the steering action of the robot is a little bit unstable when approaching an unseen obstacle (printer), it can still pass the doorway successfully. For the corridor case, an obstacle is placed in the middle of the corridor. As shown in \figurename{\ref{fig:driving through a corridor}}, the robot can navigate smoothly through the narrow space between the obstacle and the wall. Similarly, robot can be controlled safely in an office room which is a more complicated environment with many previously unseen objects in the simulator. The experiments validate that the trained D3QN model is able to enable the robot to avoid obstacles by only using a monocular camera in different real environments by benefiting from knowledge learnt in virtual environments. 

\begin{figure}
\centering
  \begin{subfigure}{0.3\linewidth}
  \centering
  \includegraphics[width=\linewidth]{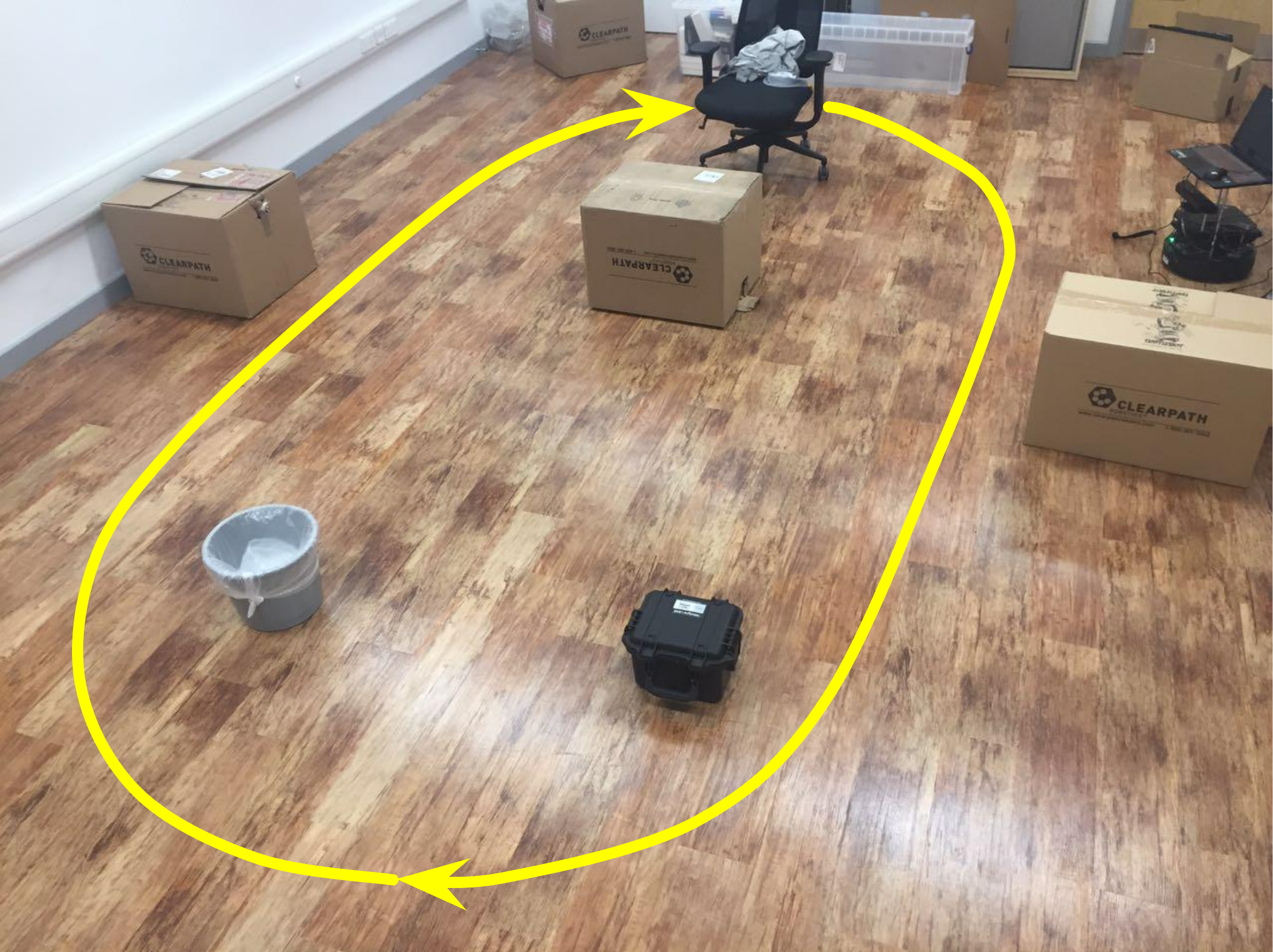}
  \label{fig:real_traj1}
  \end{subfigure}
  \vspace{-0.15in}
  \begin{subfigure}{0.3\linewidth}
  \centering
  \includegraphics[width=\linewidth]{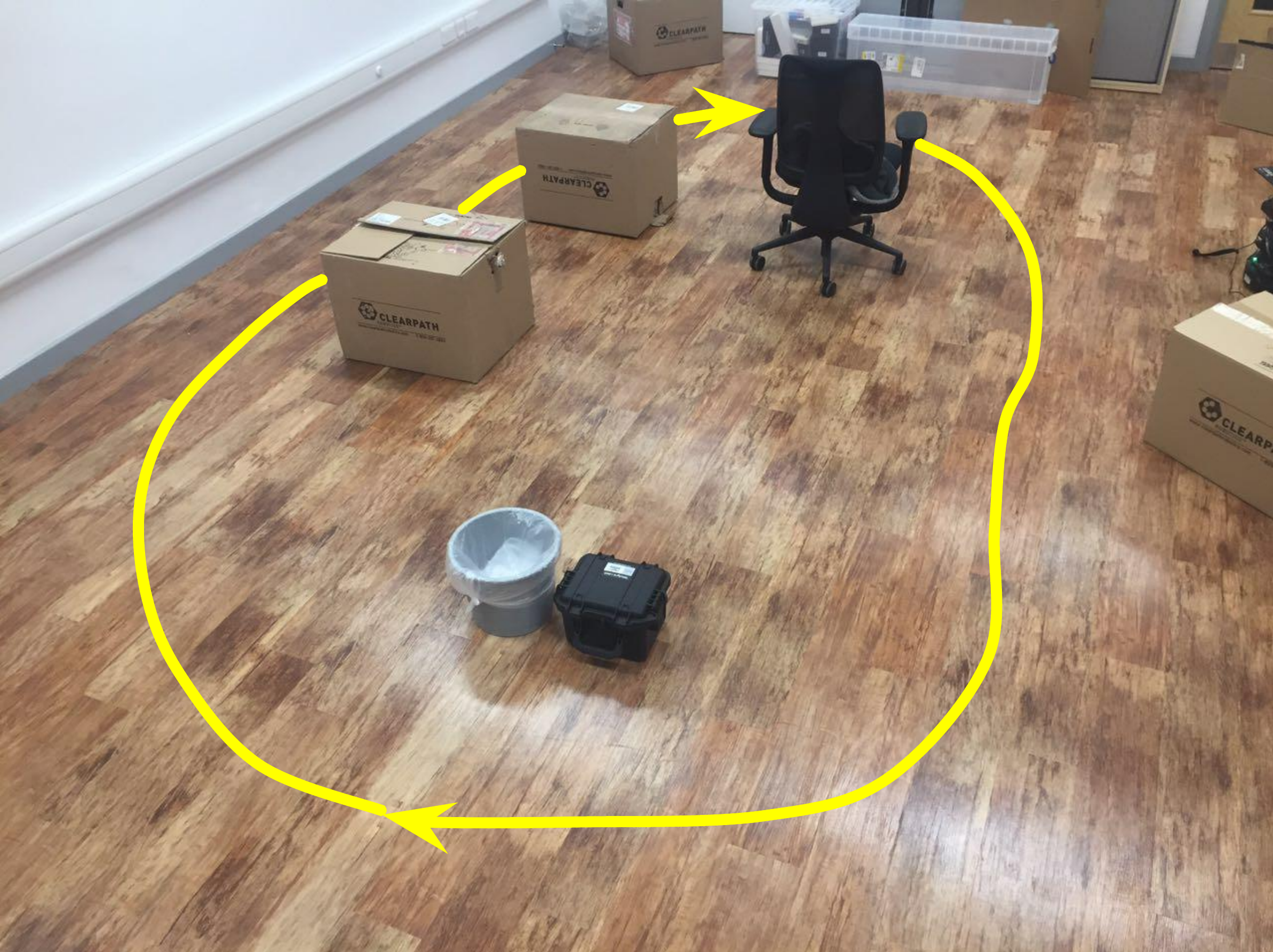}
  \label{fig:real_traj2}
  \end{subfigure}
  \begin{subfigure}{0.3\linewidth}
  \centering
  \includegraphics[width=\linewidth]{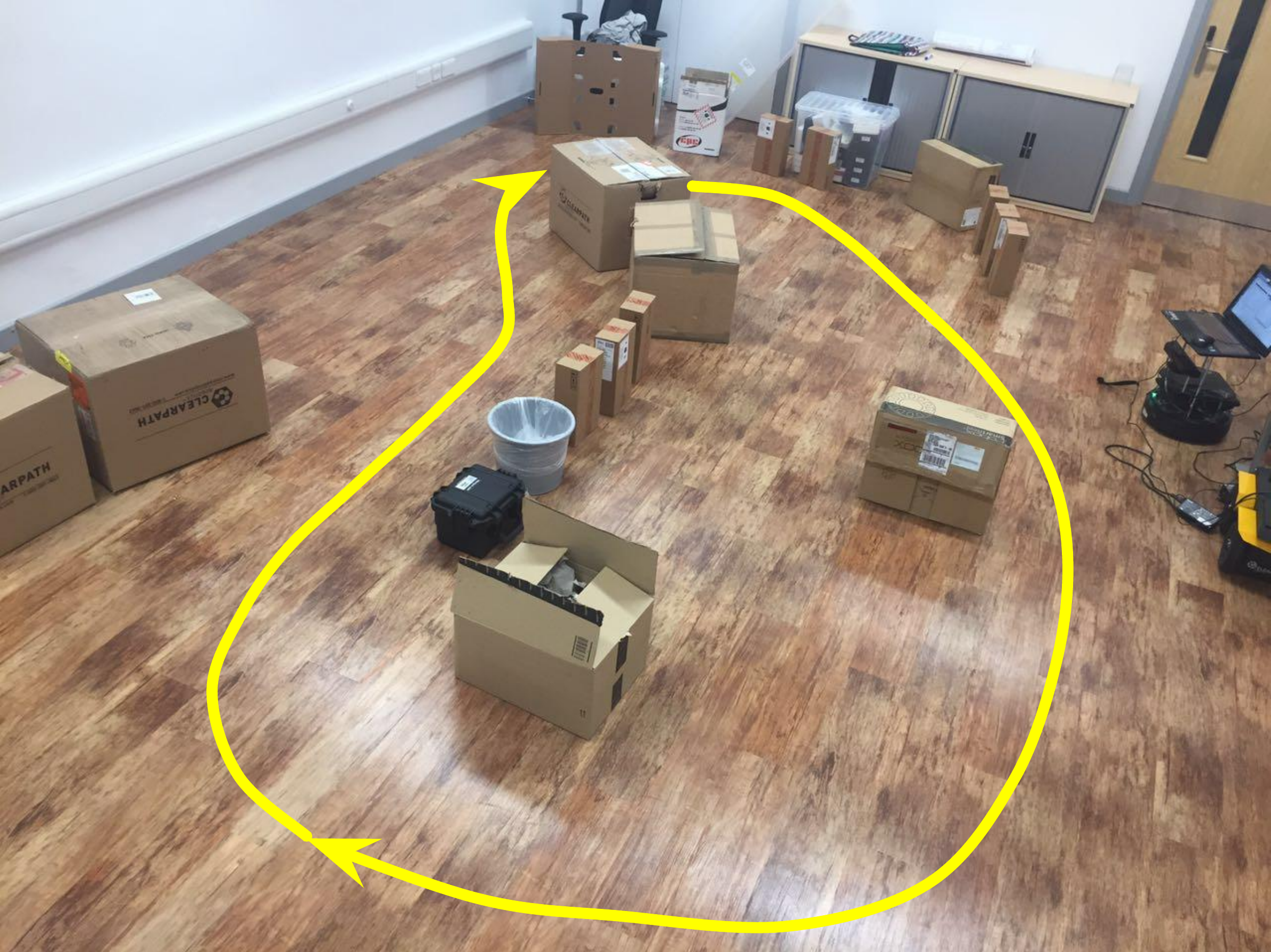}
  \label{fig:real_traj2}
  \end{subfigure}
  \begin{subfigure}{0.3\linewidth}
  \centering
  \includegraphics[angle=180, width=\linewidth]{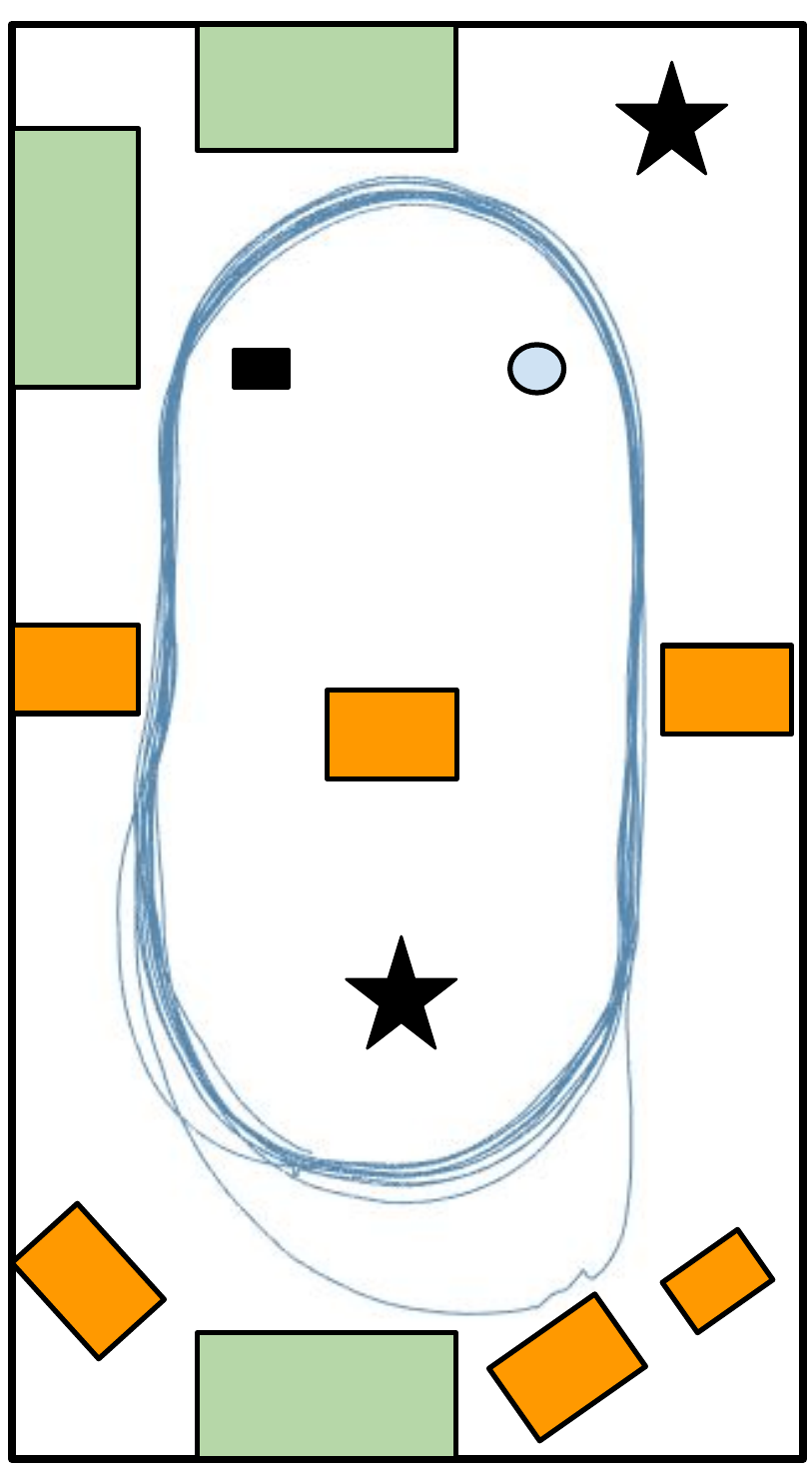}
  \caption{}
  \label{fig:traj1}
  \end{subfigure}
  \begin{subfigure}{0.305\linewidth}
  \centering
  \includegraphics[angle=180, width=\linewidth]{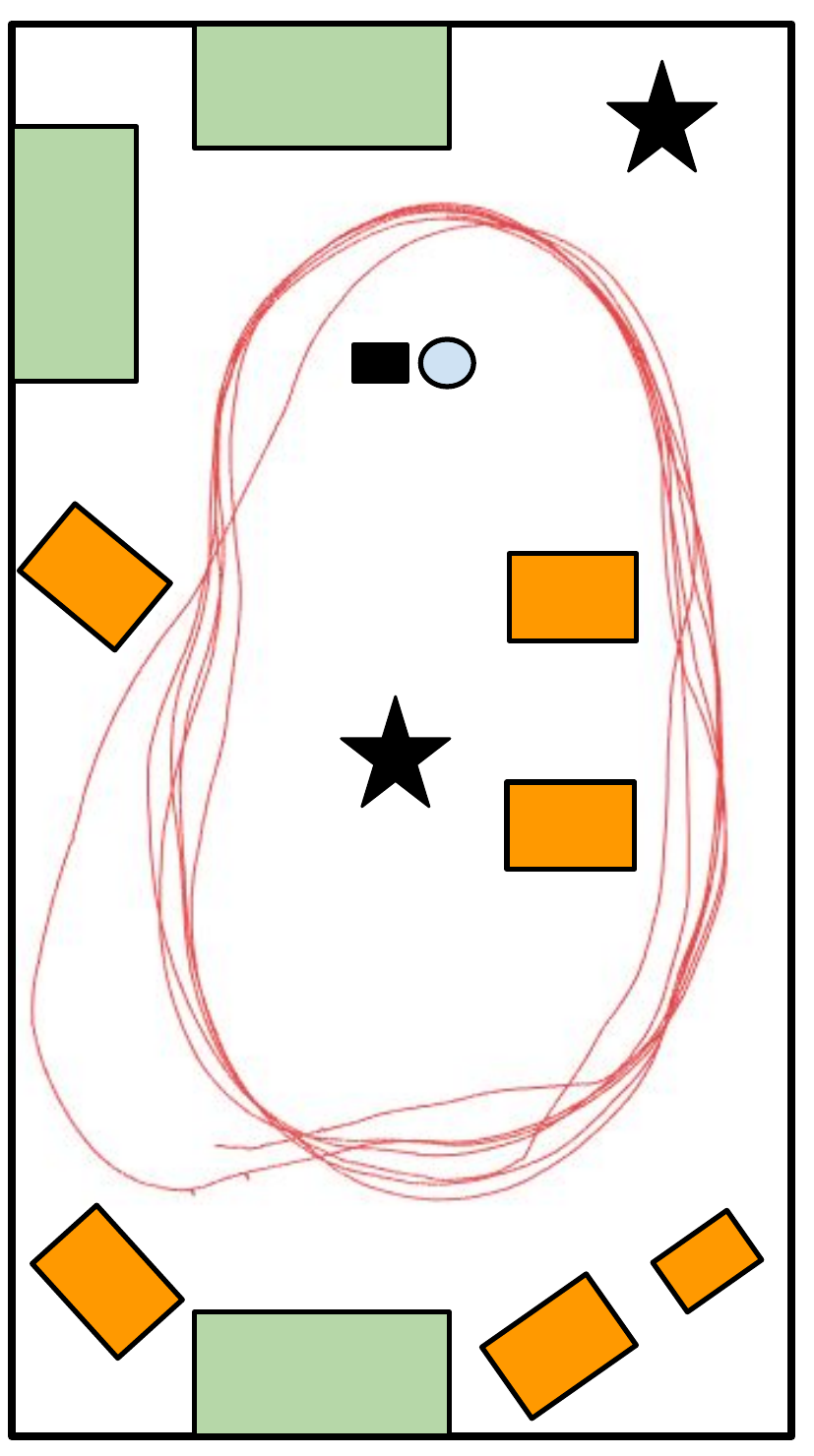}
  \caption{}
  \label{fig:traj2}
  \end{subfigure}
  \begin{subfigure}{0.3\linewidth}
  \centering
  \includegraphics[angle=180, width=\linewidth]{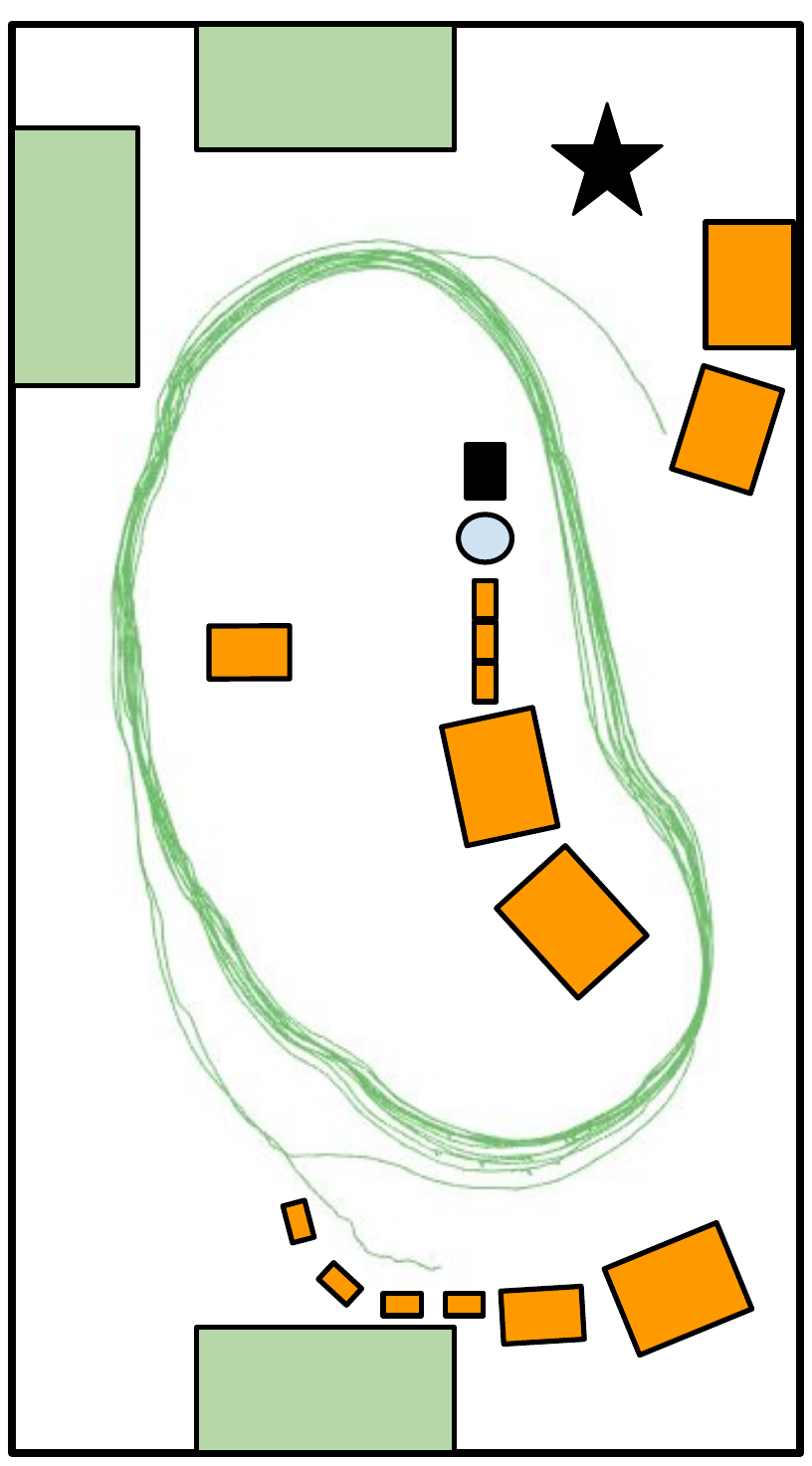}
  \caption{}
  \label{fig:traj3}
  \end{subfigure}
  \caption{Real world tests in a room with different number and placement of obstacles. Rectangles show boxes while stars and circles are chairs and trash cans respectively.}
  \label{Fig:vicon_test}
\end{figure}

\subsubsection{Tests in a Cluttered Environment}

Several long-term experiments are conducted in a cluttered room to further test the performance with dynamic layouts and objects. A Vicon system is used to record the ground truth poses of the robot. 

Fig.\ref{Fig:vicon_test} records the trajectories of the robot when it is operating around many obstacles. Green rectangles are fixed furniture while the orange ones are movable boxes. Other obstacles include two chairs (stars), a trashcan (circle) and a small suitcase (back rectangle). From the results we can see that the robot usually chooses to go along a similar path. This is because that after the Q value of each state and action pair is predicted by network, the action is selected by a greedy policy, resulting a fixed policy for all states. Since the reward function defined in the training phase prefers going in a straight line than turning, the robot navigates as a loop with the smallest curvature to maintain a maximum linear speed.

\begin{figure}
\centering
  \begin{subfigure}{0.35\linewidth}
    \centering
    \includegraphics[width=\linewidth]{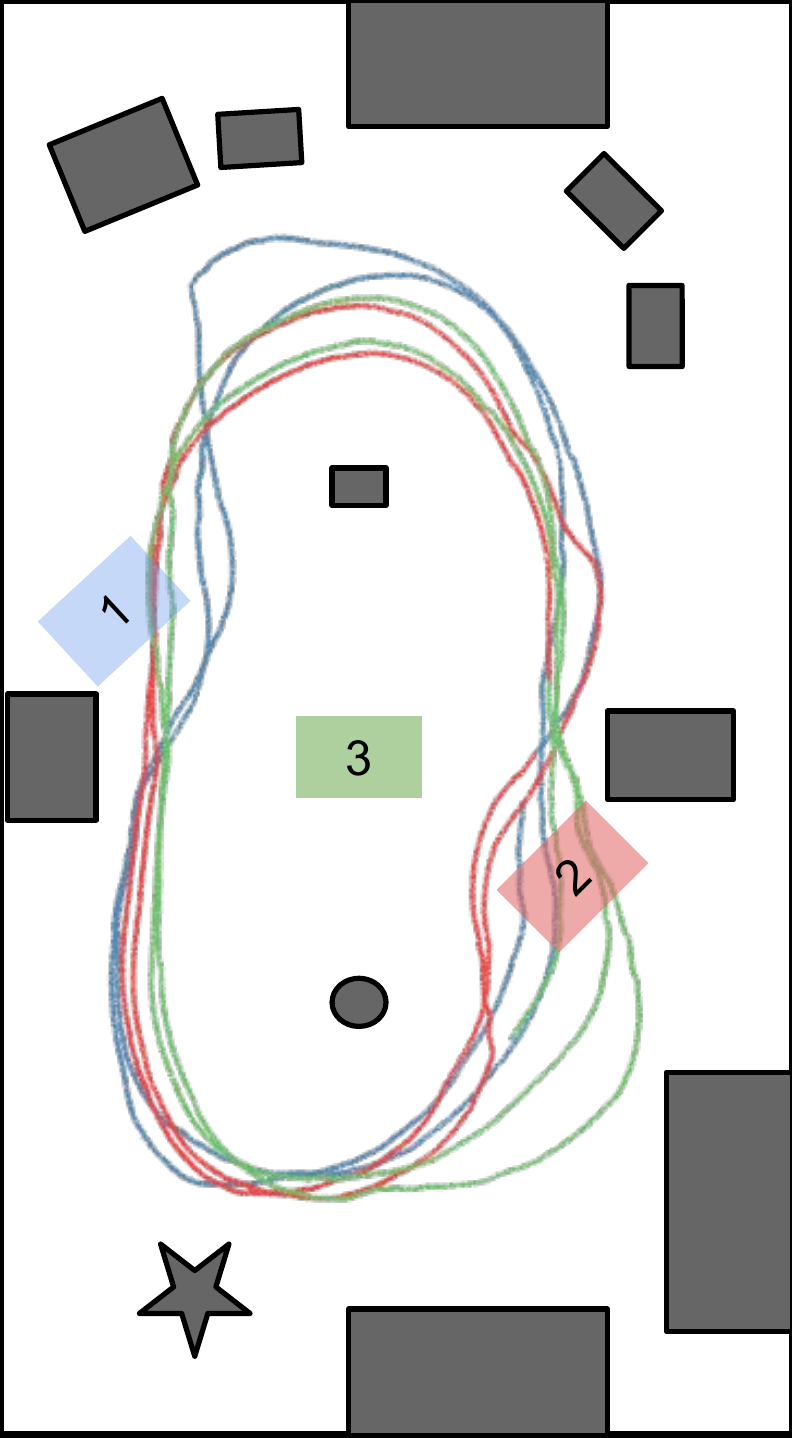}
    \label{fig: traj4}
    \caption{Simple}
  \end{subfigure}
    \begin{subfigure}{0.35\linewidth}
    \centering
    \includegraphics[width=\linewidth]{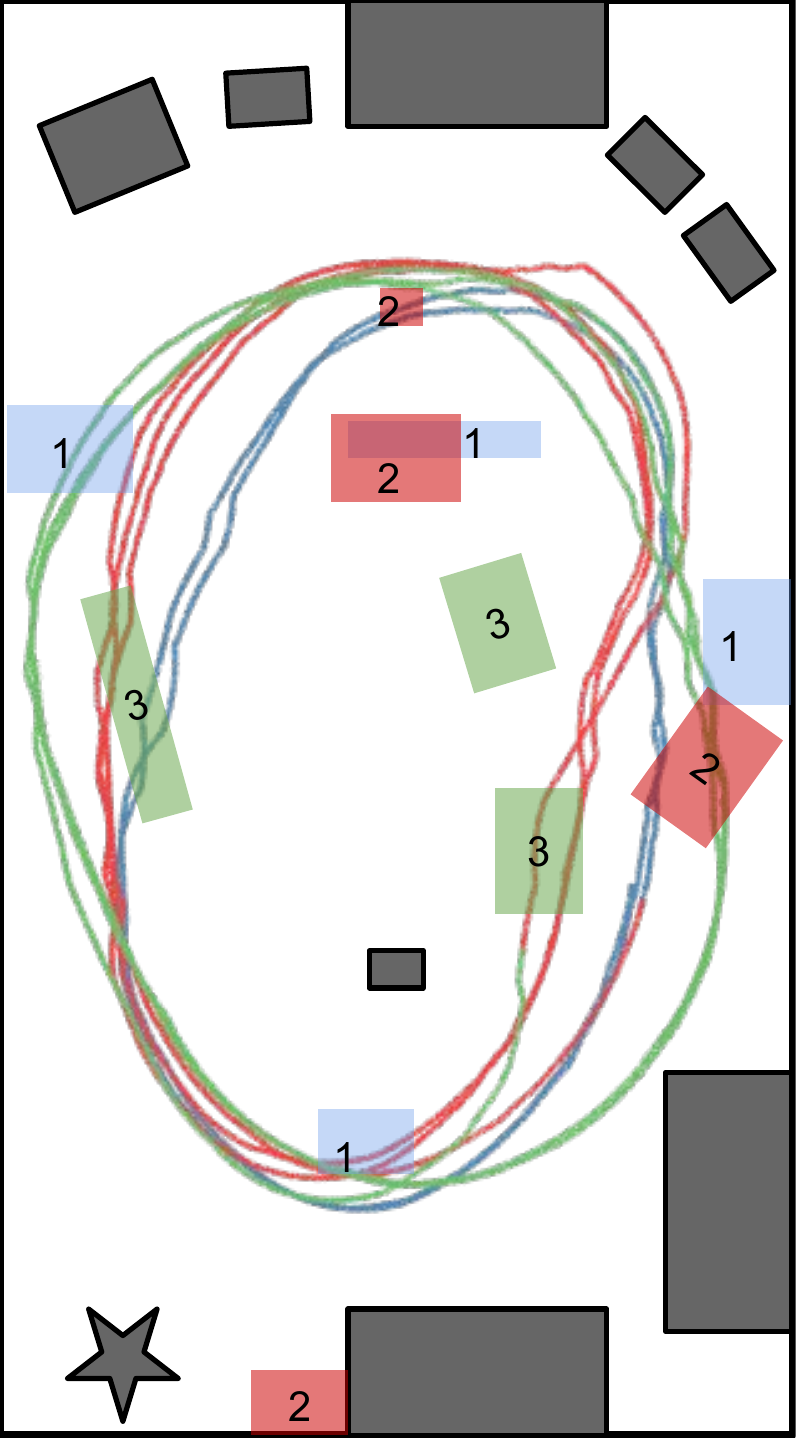}
    \label{fig: traj5}
    \caption{Complex}
  \end{subfigure}
  \caption{Real world test in dynamic environments. The obstacles in gray are fixed while colored ones indicate the motion of movable obstacles at each time. The number on the obstacles indicates the changing sequence.}
  \label{fig:dynamic test}
\end{figure}

Fig.\ref{fig:dynamic test} presents the results when the robot is tested on two dynamic environments with different complexities. Although we tried to significantly change the dynamic objects in the environments, the robot was able to avoid them by using a monocular camera, which further verifies the effectiveness of the proposed method. The video of another test is available at \url{https://youtu.be/qNIVgG4RUDM}.

\section{Conclusion} 
\label{sec:conclusion}

In this paper, a deep reinforcement learning based algorithm is proposed for obstacle avoidance by only using monocular RGB images as input. The network can be trained solely in the simulator and then directly transferred to the real-world tasks. D3QN, which is based on dueling and double network techniques, demonstrates a higher learning efficiency than normal DQN in this task and can learn from very noise depth predictions. Extensive experiments in reality demonstrate the feasibility of transferring visual knowledge of the trained network from virtual to real and the high performance of obstacle avoidance by using monocular vision.

In the future, this network will be augmented to have a more complex structure and trained with auxiliary loss functions to learn tasks such as exploration and global navigation.

\bibliographystyle{plainnat}
\bibliography{references}

\end{document}